\documentclass[10pt,twocolumn,letterpaper]{article}

\usepackage[review,algorithms]{wacv}      % To produce the REVIEW version for the algorithms track
\usepackage{graphicx}
\usepackage{amsmath}
\usepackage{amssymb}
\usepackage{booktabs}
\usepackage{mathtools}
\usepackage{multirow}
\usepackage{enumitem}

\usepackage[pagebackref,breaklinks,colorlinks]{hyperref}

\usepackage[capitalize]{cleveref}
\crefname{section}{Sec.}{Secs.}
\Crefname{section}{Section}{Sections}
\Crefname{table}{Table}{Tables}
\crefname{table}{Tab.}{Tabs.}

\def\wacvPaperID{1161} % *** Enter the WACV Paper ID here
\def\confName{WACV}
\def\confYear{2025}

\begin{document}

\title{McCaD: Multi-Contrast MRI Conditioned, Adaptive Adversarial Diffusion Model for High-Fidelity MRI Synthesis}
\author{
    Sanuwani Dayarathna\textsuperscript{1} \quad
    Kh Tohidul Islam\textsuperscript{2} \quad
    Bohan Zhuang\textsuperscript{1} \quad
    Guang Yang\textsuperscript{3}\\
    Jianfei Cai\textsuperscript{1} \quad
    Meng Law\textsuperscript{4} \quad
    Zhaolin Chen\textsuperscript{1,2}\\
    \\
    \textsuperscript{1}Department of Data Science and AI, Faculty of Information Technology, Monash University\\
    \textsuperscript{2}Monash Biomedical Imaging, Monash University\\
    \textsuperscript{3}Bioengineering Department and Imperial-X, Imperial College\\
    \textsuperscript{4}Department of Neuroscience, Monash University
}

\maketitle

%%%%%%%%% ABSTRACT
\begin{abstract}
   Magnetic Resonance Imaging (MRI) is instrumental in clinical diagnosis, offering diverse contrasts that provide comprehensive diagnostic information. However, acquiring multiple MRI contrasts is often constrained by high costs, long scanning durations, and patient discomfort. Current synthesis methods, typically focused on single-image contrasts, fall short in capturing the collective nuances across various contrasts.  Moreover, existing methods for multi-contrast MRI synthesis often fail to accurately map feature-level information across multiple imaging contrasts. We introduce McCaD (Multi-Contrast MRI Conditioned Adaptive Adversarial Diffusion), a novel framework leveraging an adversarial diffusion model conditioned on multiple contrasts for high-fidelity MRI synthesis. McCaD significantly enhances synthesis accuracy by employing a multi-scale, feature-guided mechanism, incorporating denoising and semantic encoders. An adaptive feature maximization strategy and a spatial feature-attentive loss have been introduced to capture more intrinsic features across multiple contrasts. This facilitates a precise and comprehensive feature-guided denoising process. Extensive experiments on tumor and healthy multi-contrast MRI datasets demonstrated that the McCaD outperforms state-of-the-art baselines quantitively and qualitatively. The code is provided with supplementary materials.
\end{abstract}

%%%%%%%%% BODY TEXT
\section{Introduction}
\label{sec:intro}

Magnetic Resonance Imaging (MRI) stands out as a pivotal imaging technique in clinical diagnostics, favored for its non-invasive nature. This characteristic sets it apart from other modalities ~\cite{Wang2020}. Unique to MRI is its multi-contrast capability, offering diverse information, each elucidating distinct pathological details ~\cite{Wu2010,Chen2022}. The ability to visualize various contrasts is instrumental in accurate clinical decision-making. Consequently, multi-contrast MRI has become integral in various clinical applications, including tasks like medical image segmentation and registration \cite{Zhan2022}. Despite its benefits, multi-contrast MRI faces logistical challenges, primarily in scanning time efficiency and cost-effectiveness. Additionally, the imaging process is susceptible to artifacts such as patient motion, which can compromise the quality of the acquired contrasts \cite{Dayarathna2024}. These limitations often result in varying availability of each contrast in clinical settings.

Addressing these challenges, medical image synthesis emerges as a viable approach to imputing missing data by transforming a source image modality to a desired target modality. While existing methods have predominantly focused on translating single MRI contrasts, they often neglect the rich information inherent across multiple contrasts. The advent of deep learning has brought forth advanced methods in multi-contrast MRI synthesis, outperforming traditional single-image synthesis techniques \cite{Dayarathna2024}. These approaches, mostly rooted in conventional Convolutional Neural Networks (CNNs) \cite{Chartsias2018,Osman2022,Salem2019}, have excelled in extracting informative features. However, they tend to generate suboptimal-quality images due to limitations in conventional loss functions and reliance on small neighborhood correlations. Alternatively, Generative Adversarial Network (GAN)-based methods have shown promising results \cite{Yurt2021,Sharma2019,Islam2023}, but they are often hampered by issues like mode collapse and lack of generalizability \cite{Isola2016}.

Conditional diffusion-based models have recently emerged as a promising approach, acclaimed for their ability to optimize data likelihood, thereby achieving superior sample quality and diversity \cite{Ho2020,Dhariwal2021}. However, they face specific challenges when dealing with multi-contrast MRI synthesis. \textbf{Firstly}, the complexity of capturing the intricate and complementary details across different MRI contrasts often results in synthesized images that lack the required detail and nuance for accurate diagnosis. This limitation is particularly concerning, as it may compromise the diagnostic utility of the synthesized images by failing to adequately highlight areas of clinical significance. \textbf{Secondly}, there's a critical need for these models to accurately represent essential anatomical and pathological features, a challenge compounded by their tendency to overlook crucial details in critical regions without adequate contrast-specific fine-tuning. Furthermore, the high computational demands and extended sampling times associated with diffusion-based methods pose significant constraints, particularly in medical settings where efficiency is paramount. To address these challenges, we introduce McCaD, a novel adversarial diffusion framework designed for accurate multi-contrast MRI synthesis. Our main contributions include:

\begin{itemize}[nosep]
\item A multi-scale, feature-guided mechanism that captures detailed information across various contrasts, enhancing synthesis fidelity.
\item An adaptive feature-maximization module that optimizes the synthesis process by integrating relevant features across MRI contrasts.
\item A spatial feature-attentive loss function that ensures the accurate representation of critical anatomical regions and improved pathology representation in the synthetic images.
\item Demonstrate the proposed method's effectiveness on pathological cases and healthy subjects, showcasing its superior performance over other state-of-the-art (SOTA) methods.
\end{itemize}

We have included extensive experiments on synthesizing T1-weighted (T1w), T2-weighted (T2w), and T2-fluid-attenuated inversion recovery (FLAIR) MRI contrasts from other available contrasts. We anticipate this comprehensive approach underscores the versatility and robustness of our method in various clinical scenarios.

%-------------------------------------------------------------------------

\section{Related Works}
\label{sec:literature}

Deep learning methodologies have become pivotal in medical image synthesis, adeptly capturing more complex and non-linear relationships among various image modalities \cite{Dayarathna2024}. This significant advancement has enhanced multi-contrast MRI synthesis, particularly for imputing missing or corrupted images. Generative Adversarial Networks (GANs) represent the current state-of-the-art framework for medical image synthesis tasks among more diverse applications of deep generative approaches. Their adversarial-based training has been adapted into different architectural variants such as pix2pix, pGAN, AttentionGAN, and PTNet \cite{pix2017,Dar2019, oktay2018, Zhang2021}. In addition, many novel approaches \cite{Dar2019, Uzunova2020, Mao2022} have been introduced to address the underlying contrast mapping between MRI images by characterizing the distribution of target contrast.   Moreover, multi-modal image synthesis methods \cite{Chartsias2018, Li2019, Yurt2021, Dalmaz2022} have shown more promising results than single-image-to-image translation by utilizing complementary information across multiple MRI contrasts. These approaches leverage the combined characteristics from different contrasts to achieve high-quality synthesis outcomes. For instance, MM-GAN \cite{Sharma2019} employs pre-imputation of all missing contrasts and combines them with other available contrasts for the synthesis using a curriculum-based learning approach. Hi-Net \cite{zhou2020hi} introduces a novel hybrid fusion network for multi-modal MRI synthesis by utilizing modality-specific networks to learn individual representations of contrasts and a fusion network to combine each latent representation. However, these methods are limited by less efficient fusion of multi-contrast features and often overlook the contribution of each contrast to the target modality.

 Despite the successful application of GAN-based approaches in medical image synthesis, they often exhibit distorted sampling outcomes with only pixel-level constraints. Additionally, GANs have limitations in fully capturing image diversity compared to likelihood-based models.  They can be challenging to train due to issues like mode collapse, requiring careful tuning of hyperparameters and maximizes \cite{Dayarathna2024}. Recently, likelihood-based models have garnered increased attention in high-fidelity image synthesis. Among them, Diffusion models have demonstrated remarkable image synthesis quality compared to GANs \cite{Dhariwal2021}. While diffusion-based models have been widely used in unconditional medical image generation, their application in medical image synthesis has been relatively underexplored until recently. Few recent studies \cite{Yoon2023, Pan2023, Meng2022, Zhu2023, Ozbey2022} have investigated diffusion-based approaches for medical image synthesis, employing various conditioning techniques and network enhancements. Moreover, to address the inherent drawback with diffusion models, specifically the extended sampling duration, the denoising process has been reformulated to facilitate rapid sampling by modeling the denoising distribution as an expressive multimodal distribution using GANs \cite{Xiao2021}. The first adversarial Diffusion model known as SynDiff \cite{Ozbey2022} was introduced for an unsupervised MRI synthesis by employing a cycle-consistent framework. However, most of these studies focused on single-image-to-image translation primarily due to the complex task of ensuring that the precise adjustments of the predicted posterior mean distribution from the denoising process are guided by more precise information from the conditional features across multiple contrasts considering their feature-wise contribution to the target distribution.

\section{Methodology}
\label{sec:method}

% \begin{figure*}[htb]
% \includegraphics[width=\textwidth]{Figures/Framework/mccad_architecture.pdf}
% \caption{Network architecture of McCaD. \textbf{A}: Overall Architecture, \textbf{B}: Muti-scale Feature Guided Denosing Network, \textbf{C}: Adaptive Feature Maximizer, \textbf{D}: Feature Attentive Loss.} \label{afa_diff_architecture}
% \end{figure*}

\begin{figure*}[htb]
\includegraphics[width=\textwidth]{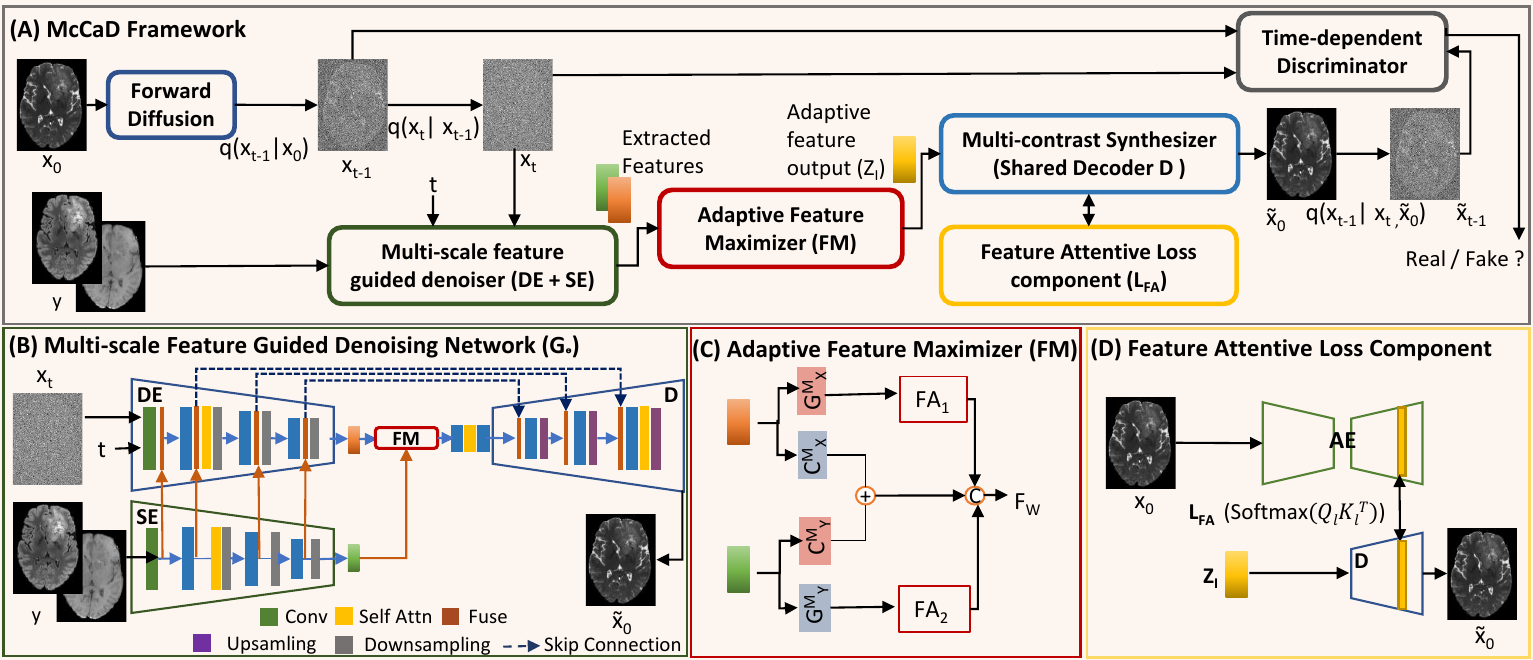}
\caption{Network architecture of McCaD. \textbf{A}: Overall Architecture, \textbf{B}: Multi-scale Feature Guided Denoising Network to incorporate feature characteristics from conditional MRI contrasts at various stages to guide the reverse diffusion process, \textbf{C}: Adaptive Feature Maximizer, to weights more pertinent features within the latent space \textbf{D}: Feature Attentive Loss to improve the perceptual quality of the synthetic results. 
} \label{afa_diff_architecture}
\end{figure*}

In response to the highlighted challenges, our methodology section delves into the details of McCaD's design. The proposed McCaD adopts a diffusion-based methodology to synthesize missing MRI contrast from multiple MR contrasts, as illustrated in Fig. \ref{afa_diff_architecture} (A), which consists of two primary phases: forward and reverse diffusion processes. In the forward diffusion process, the target MR contrast ($x$) is diffused by introducing random Gaussian noise across a series of sufficiently large time steps, resulting in a noisy image. This forms an isotropic Gaussian distribution for a large enough  time step, $T$, which can be formulated as a Markov Chain (eq. \ref{eqn:equation1}), 

\begin{equation}
 \begin{aligned}
  q(x_t | x_{t-1}) = \mathcal{N} (x_t;\sqrt{1-\beta_t}x_{t-1}, \beta_t I)
   \label{eqn:equation1}
 \end{aligned}
\end{equation}
where $\beta_t$ is the noise variance schedule used to add noise to the input data, $\mathcal{N}$ is the Gaussian distribution, and $I$ is the identity covariance matrix. The reverse diffusion process approximates the posterior distribution $p_\theta (x_{t-1}|x_t)$ to reconstruct a realistic $x_0$ from random noise (eq. \ref{eqn:equation2}).

\begin{equation}
 \begin{aligned}
p_\theta (x_{t-1}|x_t) = \mathcal{N} (x_{t-1};\mu_\theta(x_t,t),\sum_{\theta}^{}(x_t,t))
   \label{eqn:equation2}
 \end{aligned}
\end{equation}

However, as the number of steps increases, the time taken to denoise the noisy image is substantially extended, and there is no assurance that the Gaussian assumption consistently holds throughout this process. To address this challenge, we modeled the denoising distribution as a multimodal distribution with a reduced number of steps, using a GAN  to approximate the true distribution followed by an adversarial diffusion-based approach \cite{Xiao2021}.

% \begin{figure}
% \includegraphics[width=\textwidth]{McCaD.pdf}
% \caption{Network architecture of McCaD. \textbf{A}: Overall Architecture, \textbf{B}: Muti-scale Feature Guided Denosing Network, \textbf{C}: Adaptive Feature Maximizer, \textbf{D}: Feature Attentive Loss.} \label{afa_diff_architecture}
% \end{figure}

The proposed adversarial diffusion models consist of a generator, $G_\theta$, and discriminator, $D_\theta$ model for modeling the denoising diffusion process with the reduced time steps, $T=4$ for the forward diffusion process as in eq.\ref{eqn:equation1}.  We employ $G_\theta$ to approximate the distribution of $\tilde{x}_{t-1} \sim p_\theta(x_{t-1}|x_t,y)$, with remaining conditioning contrasts denoted as $y = concat(y_1,y_2)$  which acts as conditioning priors. $G_\theta$ takes the noisy target image at time step t along with the random latent vector $z$ and predicts the denoised contrast of the $x_t$, which is $\tilde{x}_0$.  Then, with the predicted  $\tilde{x}_0$, the estimated denoised step of the ${x}_{t-1}$ ($\tilde{x}_{t-1}$) can be obtained as in eq. \ref{eqn:equation3}.

\begin{equation}
 \begin{aligned}
p_\theta(x_{t-1}|x_t,y)\sim q(x_{t-1}|x_t, \tilde{x}_0 = G_\theta(x_t,y,z,t))
   \label{eqn:equation3}
 \end{aligned}
\end{equation}

To enable a multi-scale feature-guided denoising process, we have adopted a multi-encoder architecture within our denoising model, comprising a Denoising Encoder ($DE$) and a Semantic Encoder ($SE$) as in Fig. \ref{afa_diff_architecture}(B).  

\subsection{Semantic Encoder (SE)}
% The primary objective of $SE$ (Fig. \ref{afa_diff_architecture}(B)) is to generate feature representations from the $y$ across multiple scales (eq. \ref{eqn:equation4}).

%  \begin{equation}
%  \begin{aligned}
% p_\theta(x_{t-1}|x_t,y)\sim p_\theta(x_{t-1}|x_t, \phi_k (y) ))
%    \label{eqn:equation4}
%  \end{aligned}
% \end{equation}
% where $\phi_k$ gives the feature output of the corresponding ResNet block at $k^{th}$ downsampling layer in $SE$ where $k = 1,2,..,n$.

The primary objective of $SE$ (Fig. \ref{afa_diff_architecture}(B)) is to generate feature representations from the conditioning contrasts $y$ across multiple scales. The lower level feature represents the finer details within the MRI contrasts, such as textual patterns, edge details, and intensity values, and the higher scale features extract more global features, such as tissue contrasts and spatial relationships. Hence, the multi-scale feature representations are effectively utilized to condition the reverse diffusion process within the denoising network as follows (eq. \ref{eqn:equation4}).

 \begin{equation}
 \begin{aligned}
p_\theta(x_{t-1}|x_t,y)\sim p_\theta(x_{t-1}|x_t, \phi_k(y) ))
   \label{eqn:equation4}
 \end{aligned}
\end{equation}

%  \begin{equation}
%  \begin{aligned}
% \phi_k(y) = \phi_k(R(y)
%    \label{eqn:equation9}
%  \end{aligned}
% \end{equation}

\noindent $\phi_k$ is the $k^{th}$ downsampling layer in $SE$ where $k = 1,2,..,n$.

\subsection{Diffusive Encoder (DE)}
Consequently, encoder $DE$ (Fig. \ref{afa_diff_architecture}(B)) is encouraged to prioritize the contextual information received from $SE$ when predicting denoised image samples to ensure that relevant feature information is preserved in the denoising process as in eq. \ref{eqn:equation5} . These conditional features, spanning different scales, are propagated to the shared decoder $D$ through skip connections and concatenated along the channel dimension where $G_\theta$ is guided through $\phi_k(y)$.

\begin{equation}
 \begin{aligned}
DE(x_t, SE(y), t, z) = DE (x_t,\phi_k(y),t,z)
   \label{eqn:equation5}
 \end{aligned}
\end{equation}

Then $G_\theta$  models the reverse diffusion process $x_{t-1}|x_t$ with hierarchical guidance $\phi_k(y)$.

\subsection{Adaptive Feature Maximizer (FM)}

Nevertheless, while multi-stage guidance proves beneficial in extracting valuable information from features at various levels, it is more challenging to maximize the mutual information between the conditional contrasts and the target MR contrast distributions. This is mainly due to intricate dependencies between multi-contrast imaging and finding more common and mutually adaptive feature representation. To overcome this challenge, we propose an adaptive feature maximize ($FM$) within the denoising network, unifying feature distributions as shown in Fig. \ref{afa_diff_architecture}(C). 

$FM$ utilizes encoded features from $DE$ and $SE$, which undergoes separate local ($C^M_X, C^M_Y$) and global ($G^M_X, G^M_Y$) feature extraction processes, followed by feature adaptive modules ($FA_X, FA_Y$)  that weight the features based on their relevance. These feature extractor modules include convolutional layers for the local feature extraction process and an additional global average pooling layer for the global feature extraction. The feature adaptive layers consist of fully connected layers followed by SiLU and Sigmoid layers to compute the weights of the features.  Subsequently, weighted features from each encoder path are concatenated to obtain aggregated feature distribution, $Z_I$, as in eq. \ref{eqn:equation6}.

\begin{multline}
    Z_I = [G^M_X(DE(x_t,  \phi_k(y),t,z)) \\ \odot FA_X[G^M_X(DE(x_t,  \phi_k(y),t,z)] + \\ C^M_X (DE(x_t,  \phi_k(y),t,z)))] + [G^M_Y(SE(y))\\  \odot FA_Y[G^M_Y(SE(y))] + C^M_Y (SE(y)))]
    \label{eqn:equation6}
\end{multline}
% \begin{equation}
%  \begin{aligned}
% Z_I = [G^M_X(DE(x_t,  \phi_k(y),t,z)) \odot FA_X[G^M_X(DE(x_t,  \phi_k(y),t,z)] + \\ C^M_X (DE(x_t,  \phi_k(y),t,z)))] + [G^M_Y(SE(y)) \odot FA_Y[G^M_Y(SE(y))] + C^M_Y (SE(y)))]
%    \label{eqn:equation6}
%  \end{aligned}
% \end{equation}

The distinction between local and global feature contrasts derived from the denoising and conditional feature distributions aids in adaptively assigning weights to more pertinent features. This adaptive weighting facilitates the selection of mutually dependent and highly effective shared representations within the latent distribution. Consequently, these representations can be leveraged to achieve more precise denoised target contrast. 

Then, we utilized a discriminator $D_\theta$ to encourage the learned feature representation to align more closely with the distribution of the target contrast. $D_\theta$ is a time-dependent discriminator which differentiate between $x_{t-1}$ and $x_t$ by deciding if $x_{t-1}$ is a plausible denoised version of $x_t$. By obtaining discriminator $D_\theta$ it helps to guide the target distribution of the denoising image towards the actual image distribution and trained by minimizing the objective (eq.  \ref{eqn:equation7}),

\begin{multline}
    L_{D_{\theta}} = \min_\theta \sum_{t\geqslant 1}^{}\mathbb{E}_{q (x_{t} |x_0,y)}[\mathbb{E}_{q(x_{t-1}|x_t,y)} \\ [- log(D_\theta (x_{t-1},x_t,t))] \\ + \mathbb{E}_{p_\theta (x_{t-1}|x_t,y)} [-log(1-(D_\theta (\tilde{x}_{t-1},x_t,t)))]]
    \label{eqn:equation7}
\end{multline}

% \begin{equation}
%  \begin{aligned}
% L_{D_{\theta}} = \min_\theta \sum_{t\geqslant 1}^{}\mathbb{E}_{q (x_{t} |x_0,y)}[\mathbb{E}_{q(x_{t-1}|x_t,y)} [-log(D_\theta (x_{t-1},x_t,t))] \\ + \mathbb{E}_{p_\theta (x_{t-1}|x_t,y)} [-log(1-(D_\theta (\tilde{x}_{t-1},x_t,t)))]]
%    \label{eqn:equation7}
%  \end{aligned}
% \end{equation}
\begin{figure*}[!htb]
    \includegraphics[width=1\textwidth]{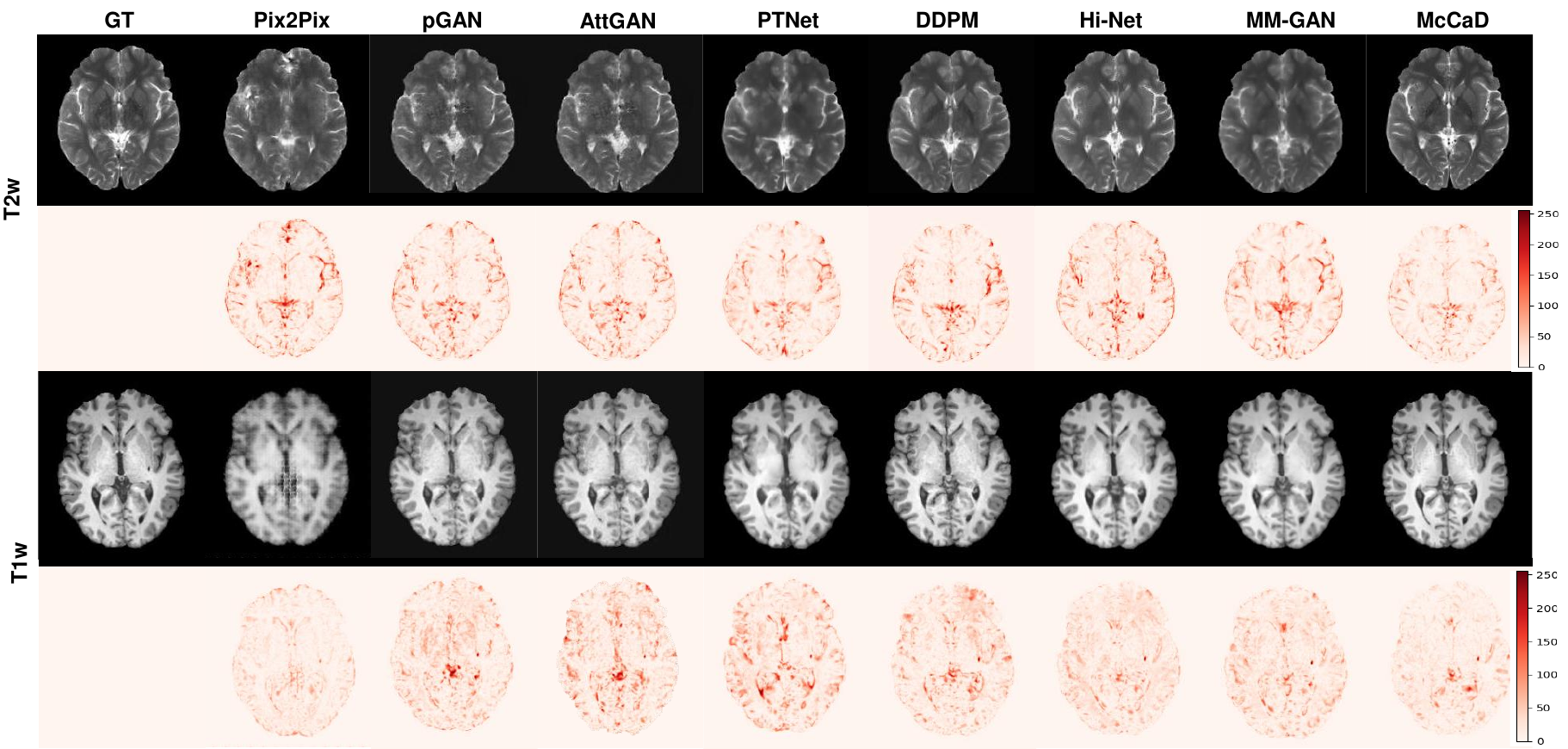}
\caption{Visualization of results for T2w and T1w Healthy brain MRI synthesis with corresponding error maps. McCaD yields lower artifacts with higher anatomical fidelity compared to baselines.} \label{qualitative_results_healthy}
\end{figure*}

\subsection{Feature Attentive (FA) Loss Function}

The self-attention mechanism \cite{vishwani2017} plays a pivotal role in diffusion models, enabling them to concentrate on more significant features within images during the denoising process. Leveraging these important attributes, we introduce a feature attentive loss function $L_{FA}$, Fig. \ref{afa_diff_architecture}(D) to encourage the model to assign higher importance to specific regions, thereby enhancing the perceptual quality of the synthesized results. To implement this, we trained an autoencoder ($AE$) with spatial attention layers in both the encoder and decoder paths with target image contrasts. 

We considered the spatial attention layer in the decoder path for feature attentive loss function, emphasizing the capture of finer details to enhance perceptual results during synthesis. For a given feature map $F_t$ $\epsilon \mathbb{R}^{(HW)\times C}$ at timestep $t$ with height $H$ and width $W$, the N-head self-attention ($A_t^h$) can be defined as in eq. \ref{eqn:equation8}, \ref{eqn:equation9}.

\begin{equation}
 \begin{aligned}
   Q_t^h = F_tW^h_Q, K_t^h = F_tW^h_K
   \label{eqn:equation8}
 \end{aligned}
\end{equation}

\begin{equation}
 \begin{aligned}
A_t^h = softmax(Q_t^h(K_t^h)^T / \sqrt{d})
   \label{eqn:equation9}
 \end{aligned}
\end{equation}

where $W^h_Q$, $W^h_K$ $\epsilon \mathbb{R}^{(C)\times d}$ and $h = 0,.., N-1$ \cite{vishwani2017}. By integrating self-attention layers into the denoising network, the model gains the capability to capture crucial features and assign distinct attention weights based on their significance. Training an autoencoder with target MRI contrasts initially results in the acquisition of attentive features through self-attention weights. Consequently, when constructing the denoising network for image synthesis, we incorporate these pre-trained self-attentive weights derived from the autoencoder's decoding process. As these weights signify essential spatial features, a feature attentive loss is computed, measuring the disparity between the attentive weights of the synthesized image decoder's attention layers and the autoencoder decoder's attention layers. This process aids the network in adjusting the target distribution of the image, potentially enhancing the quality of results, a critical aspect of diagnostic procedures. 

To calculate $L_{FA}$, we extract the attention weights of the target image contrast from the autoencoder and the diffusion model where $Q_l$, $K_l$ are features projected into queries and keys in the self-attention block \cite{vishwani2017}. These attention weights ($W^l$)  are aggregated, and global average pooling ($GAP$) is applied to merge the stacked self-attention weights from the decoder layers ($l$) (eq. \ref{eqn:equation10}). 
\vspace{0.5em}
\begin{equation}
 \begin{aligned}
    W^l =Upsample (GAP (softmax(Q_lK_l^T))
   \label{eqn:equation10}
 \end{aligned}
\end{equation}

Subsequently, the attentive loss is computed as in eq.\ref{eqn:equation11}

\begin{equation}
 \begin{aligned}
    L_{FA} = L_1 ( W^l_{Diff} - W^l_{AE})
   \label{eqn:equation11}
 \end{aligned}
\end{equation}
where $W^l_{AE}$ is the self-attention weights of target image contrast obtained from the $AE$, and $W^l_{Diff}$ is the self-attention weights obtained from the denoising process of the McCaD at corresponding layer $l$. By incorporating attentive loss,  the $G_\theta$ is trained using the non-saturating adversarial loss \cite{Goodfellow2020} and $L_1$ loss (eq. \ref{eqn:equation12}).
\vspace{1em}
\begin{multline}
    L_{G_{\theta}} = \max_\theta \sum_{t\geqslant 1}^{}\mathbb{E}_{q(x_t|x_0,y),p_\theta (x_{t-1}|x_t,y)}\\ [-log(D_\theta (\tilde{x}_{t-1},x_t,t))]
\\ +\lambda_1 L_1 (G_\theta (x_t,y,t,z), x) + \lambda_2 L_{FA}
\label{eqn:equation12}
\end{multline}
% \begin{equation}
%  \begin{aligned}
% L_{G_{\theta}} = \max_\theta \sum_{t\geqslant 1}^{}\mathbb{E}_{q(x_t|x_0,y),p_\theta (x_{t-1}|x_t,y)}[-log(D_\theta (\tilde{x}_{t-1},x_t,t))]
% \\ +\lambda_1 L_1 (G_\theta (x_t,y,t,z), x) + \lambda_2 L_{FA}
%    \label{eqn:equation10}
%  \end{aligned}
% \end{equation}
where $\lambda_1$, $\lambda_2$ is the regularization weight for $L_1$ and  $ L_{FA}$.

% \subsection{Sampling Process}

% During the inference phase, the initial denoised version of $x_t$ can be derived from the output generated by $G_\theta$ directly. However, to further preserve the stochasticity \cite{Xiao2021, Ozbey2022}, the denoised $x_t$ is obtained through a sampling process following $T$ denoising steps as follows. 

% \begin{equation}
%  \begin{aligned}
% p_\theta(x_{t-1}|x_t,y)\sim q(x_{t-1}|x_t, \tilde{x}_0 = G_\theta^T(x_t,y,z,t))
%    \label{eqn:equation20}
%  \end{aligned}
% \end{equation}

% The sampling process begins at timestep $T$ and iteratively proceeds in reverse, performing denoising steps according to eq. \ref{eqn:equation20} or a total of $T$ steps.
\begin{figure*}[!htb]
    \includegraphics[width=1\textwidth]{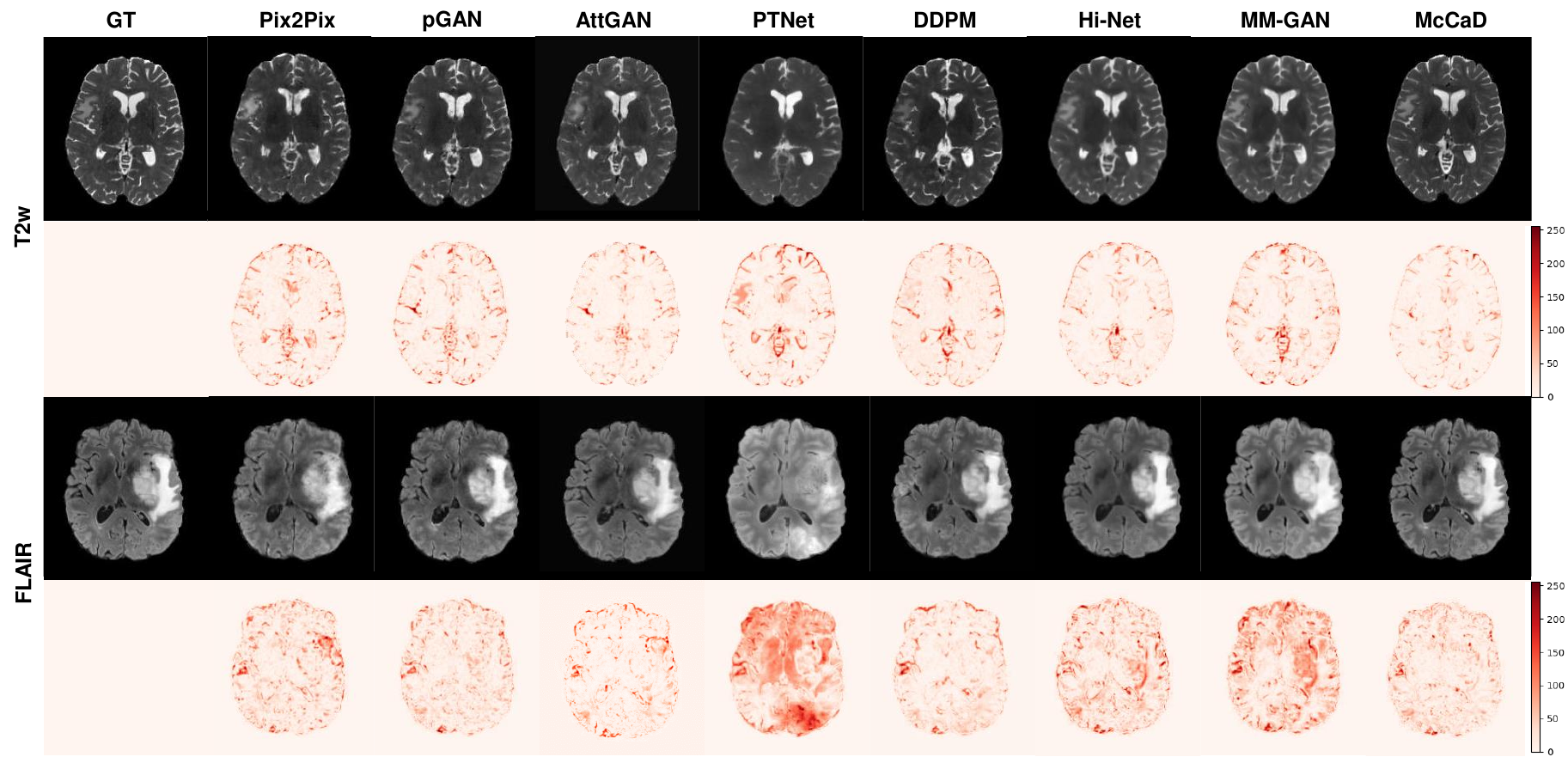}
\caption{Visualization of results for T2w and FLAIR tumor brain MRI synthesis on BraTS dataset with corresponding error maps. McCaD shows improved synthetic results compared to baselines with fewer errors and more accurate depictions of pathological regions.} \label{qualitative_results_brats}
\end{figure*}

\subsection{Model Architecture}

The structure of $G_\theta$ adheres to the U-Net framework \cite{Xiao2021}, incorporating six ResNet blocks in the encoder-decoder path with a latent dimension of 256. An attention layer is incorporated in the encoder and decoder paths at a scale of 16. It also integrates sinusoidal positioning embeddings for time-step conditioning followed by \cite{Ho2020}. The two multi-encoder paths ($SE$ and $DE$) share the same architecture: ResNet Blocks, followed by downsampling and feature fusioning layers. The encoded features from the separate encoders go through the Adaptive Feature maximizer module, where the weighted and fused features are directed to the shared decoder. The decoder consists of 6 ResNet blocks followed by upsampling and self-attention layers. The fused features from the $SE$ and $DE$ are concatenated in the decoder through the channel dimension through the skip connection.

The autoencoder ($AE$) replicates the architecture of $G_\theta$, adopting a similar encoder-decoder structure that includes attention blocks at scale 16.  Meanwhile, $D_\theta$ is structured as a convolutional network with ResNet components \cite{Xiao2021}, operating as a time-dependent network conditioned on $x_t$ and $t$. Time conditioning is applied using the same sinusoidal positional embeddings as in $G_\theta$, and the discriminator is provided with both the actual and denoised image sample at time $t-1$ the noisy image at time $t$, along with the corresponding time step $t$.
%-------------------------------------------------------------------------
\subsection{Experiments and Results}

\subsubsection{Datasets and Implementation Details}
We evaluated our proposed methods on two muti-contrast datasets: the Brain Tumor Segmentation Challenge (BraTS) dataset \cite{brats2021} and a private dataset for healthy subjects.  In the BraTS dataset, we randomly selected 60 subjects, with 25, 15, and 20 for training, validation, and testing.  We used 100 middle axial slices from each subject reshaped to 256 $\times$  256. The healthy dataset was obtained through the experiment conducted at Monash Biomedical Imaging using the Siemens Biograph mMR 3T MRI. Institutional ethics and review board (IRB) approvals were obtained from the Monash University Human Research Ethics Committee, and written consent from all participants involved in the study. The SynthSeg method \cite{Billot2022} was used for resampling data to $1 mm^3$ isotropic resolution and FSL FAST for the bias field correction and coregistered each contrast using FMRIB’s linear image registration tool (FLIRT) \cite{Jenkinson2002}.  For the study, we have selected 45 subjects, with 20, 10, and 15 for training, validation, and testing. From each subject, we have used 100 middle axial slices from 160 total slices reshaped to 256 $\times$ 256.

The hyperparameters for the proposed network consist of the Adam optimizer ($\beta_1$=0.5, $\beta_2$=0.9), a learning rate of 0.001, four diffusive steps, and a noise variance range from $\beta_{min}$=0.1 to $\beta_{max}$=20 \cite{Ozbey2022}. For each contrast synthesis, the network is trained for 50 epochs, and $AE$ is trained for 40 epochs in PyTorch on a single NVIDIA A40 graphics processing unit with 46 GB of memory.

% \begin{table}[ht]
% \resizebox{\columnwidth}{!}{%
% \begin{tabular}{@{}ccccccc@{}}
% \toprule
%  & \multicolumn{2}{c}{\textbf{T1w, T2w→ FLAIR}} & \multicolumn{2}{c}{\textbf{T1w, FLAIR → T2w}} & \multicolumn{2}{c}{\textbf{FLAIR, T2w → T1w}} \\ \midrule
%  & PSNR(dB) ↑ & SSIM ↑ & PSNR(dB) ↑ & SSIM ↑ & PSNR(dB) ↑ & SSIM ↑ \\ \midrule
% pix2pix & 24.56±2.28 & 0.891±0.029 & 25.19±1.58 & 0.889±0.029 & 21.73±3.02 & 0.824±0.047 \\ \midrule
% pGAN & 25.66±2.29 & 0.896±0.029 & 25.77±1.58 & 0.896±0.025 & 21.87±3.37 & 0.859±0.034 \\ \midrule
% AttGAN & 25.92±2.32 & 0.869±0.057 & 26.26±1.72 & 0.898±0.029 & 21.79±3.43 & 0.800±0.108 \\ \midrule
% PTNet & 20.25±2.99 & 0.857±0.041 & 23.83±1.87 & 0.872±0.036 & 18.46±4.60 & 0.856±0.053 \\ \midrule
% DDPM & 24.28±2.39 & 0.796±0.061 & 23.91±1.85 & 0.876±0.033 & 21.96±3.85 & 0.881±0.049 \\ \midrule
% MM-GAN & 25.49±2.78 & 0.916± 0.028 & 26.11±2.15 & 0.914±0.024 & 23.22±4.77 & 0.920± 0.033 \\ \midrule
% Hi-Net & 25.02±2.59 & 0.915±0.026 & 25.96±1.98 & 0.918±0.026 & 21.76±4.30 & 0.908±0.034 \\ \midrule
% \textbf{McCaD} & \textbf{26.11±2.66} & \textbf{0.920±0.029} & \textbf{27.03±1.82} & \textbf{0.923±0.023} & \textbf{23.24±4.44} & \textbf{0.923±0.032} \\ \bottomrule
% \end{tabular}%
% }
% \caption{Performance comparison for BraTS dataset (mean±std.).}
% \label{tab:brats_results}
% \end{table}

\begin{table}[htb]
\resizebox{\columnwidth}{!}{%
\begin{tabular}{@{}ccccccc@{}}
\toprule
 & \multicolumn{2}{c}{\textbf{T1w, T2w→ FLAIR}} & \multicolumn{2}{c}{\textbf{T1w, FLAIR → T2w}} & \multicolumn{2}{c}{\textbf{FLAIR, T2w → T1w}} \\ \midrule
 & PSNR(dB) ↑ & SSIM ↑ & PSNR(dB) ↑ & SSIM ↑ & PSNR(dB) ↑ & SSIM ↑ \\ \midrule
pix2pix & 26.30±2.00 & 0.911±0.038 & 25.35±1.39 & 0.879±0.041 & 22.07±0.60 & 0.810±0.020 \\  \midrule
pGAN & 26.94±1.54 & 0.922±0.023 & 26.54±1.27 & 0.901±0.03 & 28.22±2.46 & 0.940±0.024 \\  \midrule
AttGAN & 26.54±1.85 & 0.913±0.053 & 25.62±1.42 & 0.883±0.042 & 28.32±2.18 & 0.904±0.088 \\ \midrule
PTNet & 26.15±2.06 & 0.921±0.041 & 25.63±1.47 & 0.900±0.034 & 25.47±2.45 & 0.919±0.074 \\ \midrule
DDPM & 26.73±2.17 & 0.812±0.063 & 25.99±1.36 & 0.529±0.057 & 26.47±2.95 & 0.924±0.028 \\ \midrule
MM-GAN & 28.25±1.86 & 0.939±0.023 & 26.33±1.59 & 0.907± 0.040 & 28.38±3.30 & 0.952±0.033 \\ \midrule
Hi-Net & 27.61±1.78 & 0.939±0.032 & 27.05±1.17 & 0.915±0.037 & 27.84±3.11 & 0.952±0.027 \\ \midrule
\textbf{McCaD} & \textbf{28.87±1.85} & \textbf{0.945±0.022} & \textbf{27.55±1.67} & \textbf{0.927±0.027} & \textbf{28.84±2.93} & \textbf{0.956±0.022} \\ \bottomrule
\end{tabular}%
}
\caption{Performance comparison for healthy dataset (mean±std.).}
\label{tab:healthy_results}
\end{table}

\begin{table}[htb]
\resizebox{\columnwidth}{!}{%
\begin{tabular}{@{}ccccccc@{}}
\toprule
 & \multicolumn{2}{c}{\textbf{T1w, T2w→ FLAIR}} & \multicolumn{2}{c}{\textbf{T1w, FLAIR → T2w}} & \multicolumn{2}{c}{\textbf{FLAIR, T2w → T1w}} \\ \midrule
 & PSNR(dB) ↑ & SSIM ↑ & PSNR(dB) ↑ & SSIM ↑ & PSNR(dB) ↑ & SSIM ↑ \\ \midrule
pix2pix & 24.56±2.28 & 0.891±0.029 & 25.19±1.58 & 0.889±0.029 & 21.73±3.02 & 0.824±0.047 \\ \midrule
pGAN & 25.66±2.29 & 0.896±0.029 & 25.77±1.58 & 0.896±0.025 & 21.87±3.37 & 0.859±0.034 \\ \midrule
AttGAN & 25.92±2.32 & 0.869±0.057 & 26.26±1.72 & 0.898±0.029 & 21.79±3.43 & 0.800±0.108 \\ \midrule
PTNet & 20.25±2.99 & 0.857±0.041 & 23.83±1.87 & 0.872±0.036 & 18.46±4.60 & 0.856±0.053 \\ \midrule
DDPM & 24.28±2.39 & 0.796±0.061 & 23.91±1.85 & 0.876±0.033 & 21.96±3.85 & 0.881±0.049 \\ \midrule
MM-GAN & 25.49±2.78 & 0.916± 0.028 & 26.11±2.15 & 0.914±0.024 & 23.22±4.77 & 0.920± 0.033 \\ \midrule
Hi-Net & 25.02±2.59 & 0.915±0.026 & 25.96±1.98 & 0.918±0.026 & 21.76±4.30 & 0.908±0.034 \\ \midrule
\textbf{McCaD} & \textbf{26.11±2.66} & \textbf{0.920±0.029} & \textbf{27.03±1.82} & \textbf{0.923±0.023} & \textbf{23.24±4.44} & \textbf{0.923±0.032} \\ \bottomrule
\end{tabular}%
}
\caption{Performance comparison for BraTS dataset (mean±std.).}
\label{tab:brats_results}
\end{table}

% \begin{table}[ht]
% \resizebox{\columnwidth}{!}{%
% \begin{tabular}{@{}ccccccc@{}}
% \toprule
%  & \multicolumn{2}{c}{\textbf{T1w, T2w→ FLAIR}} & \multicolumn{2}{c}{\textbf{T1w, FLAIR → T2w}} & \multicolumn{2}{c}{\textbf{FLAIR, T2w → T1w}} \\ \midrule
%  & PSNR(dB) ↑ & SSIM ↑ & PSNR(dB) ↑ & SSIM ↑ & PSNR(dB) ↑ & SSIM ↑ \\ \midrule
% pix2pix & 26.30±2.00 & 0.911±0.038 & 25.35±1.39 & 0.879±0.041 & 22.07±0.60 & 0.810±0.020 \\ \midrule
% pGAN & 26.94±1.54 & 0.922±0.023 & 26.54±1.27 & 0.901±0.03 & 28.22±2.46 & 0.940±0.024 \\ \midrule
% AttGAN & 26.54±1.85 & 0.913±0.053 & 25.62±1.42 & 0.883±0.042 & 28.32±2.18 & 0.904±0.088 \\ \midrule
% PTNet & 26.15±2.06 & 0.921±0.041 & 25.63±1.47 & 0.900±0.034 & 25.47±2.45 & 0.919±0.074 \\ \midrule
% DDPM & 26.73±2.17 & 0.812±0.063 & 25.99±1.36 & 0.529±0.057 & 26.47±2.95 & 0.924±0.028 \\ \midrule
% MM-GAN & 28.25±1.86 & 0.939±0.023 & 26.33±1.599 & 0.907± 0.040 & 28.38±3.30 & 0.952±0.033 \\ \midrule
% Hi-Net & 27.61±1.78 & 0.939±0.032 & 27.05±1.17 & 0.915±0.037 & 27.84±3.11 & 0.952±0.027 \\ \midrule
% \textbf{McCaD} & \textbf{28.87±1.85} & \textbf{0.945±0.022} & \textbf{27.55±1.67} & \textbf{0.927±0.027} & \textbf{28.84±2.93} & \textbf{0.956±0.022} \\ \bottomrule
% \end{tabular}%
% }
% \caption{Performance comparison for healthy dataset (mean±std.).}
% \label{tab:healthy_results}
% \end{table}

\subsubsection{Competing Methods and Evaluation Metrics}
We assessed the performance of our model by comparing it with different competing methods, including conventional generative frameworks for image synthesis such as pix2pix\cite{pix2017}, pGAN \cite{Dar2019}, Attention GAN (AttGAN) \cite{oktay2018} and transformer-based PTNet \cite{Zhang2021}, SOTA for Multi-Contrast MRI synthesis including MM-GAN \cite{Sharma2019}, Hi-Net \cite{zhou2020hi}, and conventional diffusion model DDPM \cite{Ho2020}. All methods are evaluated using structural similarity index (SSIM) and peak signal-to-noise ratio (PSNR) metrics \cite{Wang2004}. For ablation studies, we utilize Frechet Inception Distance (FID) to quantify the perceptual similarity among the images.

\subsubsection{Performance Comparison}

The qualitative comparison for McCaD and competing methods are shown in Fig. \ref{qualitative_results_healthy} and \ref{qualitative_results_brats}. The quantitative comparison of the results is presented in Tables \ref{tab:healthy_results} and \ref{tab:brats_results}. When comparing the performance of the proposed McCaD on a healthy dataset (Table \ref{tab:healthy_results}), the quantitative results highlight its superior accuracy across all synthesis tasks. McCaD outperforms conditional GAN-based approaches by up to 6.77 dB in PSNR and 14.6\% in SSIM. Compared to DDPM, McCaD shows improvements of up to 2.37 dB in PSNR and 39.8\% in SSIM. While SOTA methods for Multi-Contrast MRI synthesis, such as MM-GAN and Hi-Net, have shown competitive performance, McCaD surpasses all these methods, achieving the highest synthesis performance. The quantitative results for the tumor dataset (Table \ref{tab:brats_results}) also demonstrate that McCaD achieved the highest scores for both PSNR and SSIM across all synthesis tasks. Specifically, it surpassed GAN-based approaches by up to 1.84 dB in PSNR and 12.3\% in SSIM, and transformer-based approaches by up to 5.86 dB in PSNR and 6.7\% in SSIM.  Despite the competing results obtained by Multi-Contrast MRI synthesis methods for tumor synthesis, McCaD outperformed these methods in more challenging scenarios.

Examining each synthesis task individually, McCAD significantly enhances the qualitative aspects of synthesis images with minimal error in both healthy and pathological cases. From visual inspection of healthy synthesis results (Fig. \ref{qualitative_results_healthy}), McCaD demonstrates superior image details compared to other methods with an accurate synthesis of ventricles (T2w), cortical and subcortical structures (T1w). Similarly, the qualitative representation of tumor synthesis results (Fig. \ref{qualitative_results_brats}) shows more accurate tumor contrasts from McCaD with precise boundary information in both T2w and FLAIR images compared to other methods. While AttGAN and DDPM produce better results, they tend to introduce blurriness in the synthesis contrasts, diminishing the perceptual quality of the images. PTNet has demonstrated less precise outcomes due to its heavy dependence on a self-attention mechanism without the inclusion of convolutional layers, thereby constraining its capability for accurate localization of features. In comparison to SOTA Multi-Contrast MRI synthesis, McCaD has shown to excel in capturing more precise structural details of the brain tissues with more clarity by preserving the sharpness of the boundaries in the lesion regions. Therefore, the results undeniably highlight the superior quality synthesis of images that closely resemble ground truths in both cases.

\subsubsection{Tumor Segmentation}

% \begin{table}[htb]
% \resizebox{\columnwidth}{!}{%
% \begin{tabular}{@{}ccccccc@{}}
% \toprule
%  & \textbf{AttGAN} & \textbf{DDPM} & \textbf{Hi-Net} & \textbf{MM-GAN} & \textbf{McCaD} & \textbf{Complete} \\ \midrule
 
% T2w & 0.7692 & 0.7987 & 0.7921 & 0.7991 & \textbf{0.8004} & 0.8188 \\
% T2w + FLAIR & 0.6974 & 0.6262 & 0.7401 & 0.7449 & \textbf{0.7602} & 0.8188 \\ \bottomrule
% \end{tabular}%
% }
% \caption{Quantitative evaluation of downstream segmentation task.}
% \label{tab:seg-results}
% \end{table}

\begin{figure}[!ht]
    \includegraphics[width=\columnwidth]{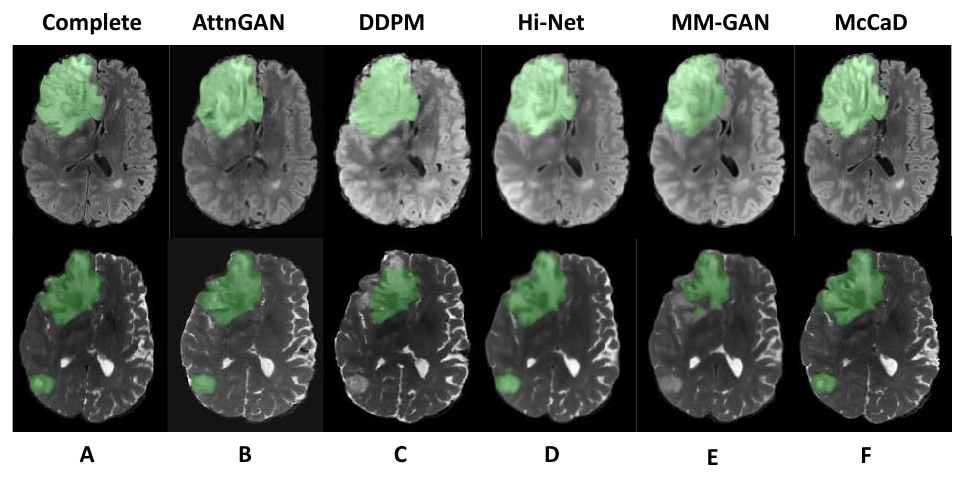}
    \caption{Visualization of tumor segmentation results. Column (A) shows the tumor mask segmented using ground truths. Using different methods, the first row (B-F) shows the tumor segmentation masks with T1w, synthetic FLAIR, and synthetic T2w. The second row (B-F) shows tumor segmentation masks from T1w, FLAIR, and synthetic T2w.} \label{Figure_4:segmentation_results}
\end{figure}

To demonstrate the effectiveness and diagnostic equivalence of our synthetic results, we performed tumor segmentation using synthesized contrasts on the BraTS dataset. For comparison, we selected top-performing methods in each category from the baselines. We trained a multi-modal U-Net \cite{cardoso2022monai} using the same BraTS data split that takes T1w, T2w, and FLAIR images as inputs to generate a tumor mask for the whole tumor. During testing, we initially evaluated the segmentation accuracy using all three ground truth contrasts, representing a 'complete' scenario. Then, we replaced the T2w and FLAIR contrasts with imputed images. The results in Table \ref{tab:seg-results} show the highest dice score values for segmentation maps generated using synthetic images from McCaD compared to other methods, indicating a more clinically reliable synthesis. The results for T2w synthetic segmentation accuracy demonstrate that McCaD closely resembles the actual predicted mask with precise tumor synthetic contrasts, achieving a dice score very similar to that with complete real contrasts. Replacing FLAIR resulted in slight drops in segmentation accuracy but still outperformed all other competing methods. The quantitative and visual representation of these results in Fig. \ref{Figure_4:segmentation_results} further indicates evident superiority over other baselines by achieving similar segmentation results when using ground truths as inputs.

\begin{table}[htb]
\resizebox{\columnwidth}{!}{%
\begin{tabular}{@{}ccccccc@{}}
\toprule
 & \textbf{AttGAN} & \textbf{DDPM} & \textbf{Hi-Net} & \textbf{MM-GAN} & \textbf{McCaD} & \textbf{Complete} \\ \midrule
 
T2w & 0.7692 & 0.7987 & 0.7921 & 0.7991 & \textbf{0.8004} & 0.8188 \\  \midrule
T2w + FLAIR & 0.6974 & 0.6262 & 0.7401 & 0.7449 & \textbf{0.7602} & 0.8188 \\ \bottomrule
\end{tabular}%
}
\caption{Quantitative evaluation with dice scores for downstream segmentation task.}
\label{tab:seg-results}
\end{table}

% \begin{figure*}
% \includegraphics[width=\textwidth]{Figures/Eval/ssim_distribution.pdf}
% \caption{Distribution of SSIM values across different synthesis methods for (A). T2w, (B) FLAIR synthesis from the healthy dataset and (C) T2w synthesis from the tumor dataset.} \label{afa_diff_architecture}
% \end{figure*}

\subsubsection{Ablation Studies}

We conducted a series of experiments on the BraTS dataset to showcase the significance of the key components introduced in the McCaD.  

% Table \ref{tab:ablation_results} illustrates their effectiveness in improving feature representations and Table \ref{tab:perceptual_quality_results} demonstrates the impact of the adaptive FM and FA loss on the perceptual quality. 

\textbf{Effect of multi-scale feature guidance:} 
We assessed the impact of multi-scale guidance from the conditional multi-contrast by eliminating the intermediate feature fusion layers in the denoising encoder. Thus, the guidance at various scales that flow through the shared decoder via skip connections was removed.
As shown in Table \ref{tab:ablation_results}, the results highlight the significant contribution of multi-scale feature guidance, where removing this component resulted in a drastic loss of performance in the synthesized results.

\begin{table}[]
\centering
\resizebox{\columnwidth}{!}{%
\begin{tabular}{@{}ccccccc@{}}
\cmidrule(l){2-7}
\multirow{2}{*}{} & \multicolumn{2}{c}{\textbf{T1w, Tw→ FLAIR}} & \multicolumn{2}{c}{\textbf{T1w, FLAIR → T2w}} & \multicolumn{2}{c}{\textbf{FLAIR, T2w → T1w}} \\ \cmidrule(l){2-7} 
 & PSNR & SSIM & PSNR & SSIM & PSNR & SSIM \\ \midrule
\begin{tabular}[c]{@{}c@{}}w/o multi-scale \\ feature \\ guidance\end{tabular} & 17.32±1.67 & 0.745 ± 0.048 & 20.57 ± 1.61 & 0.774 ±0.041 & 19.77 ± 2.54 & 0.810 ± 0.039 \\ \midrule
\begin{tabular}[c]{@{}c@{}}w/o adap. FM \\ and FA loss\end{tabular} & 25.89±2.45 & 0.919±0.028 & 26.73±2.021 & 0.922±0.025 & \textbf{24.31±3.99} & 0.920±0.032 \\ \midrule
McCaD & \textbf{26.11±2.66} & \textbf{0.920±0.029} & \textbf{27.03±1.82} & \textbf{0.923±0.023} & 23.24±4.44 & \textbf{0.923±0.032} \\ \bottomrule
\end{tabular}%
}
\caption{Ablation studies for multi-scale feature guidance, adaptive FM, and FA loss.}
\label{tab:ablation_results}
\end{table}

\begin{table}[]
\centering
\resizebox{\columnwidth}{!}{%
\begin{tabular}{@{}cccc@{}}
\toprule
 & \textbf{T1w, T2w → FLAIR} & \textbf{T1w, T2w → T2w} & \textbf{FLAIR, T2w → T1w} \\ \midrule
\begin{tabular}[c]{@{}c@{}}w/o adap. FM \\ and w/o FA loss\end{tabular} & 21.436 & 15.976 & 54.935 \\  \midrule
\begin{tabular}[c]{@{}c@{}}with adap. FM \\ and w/o FA loss\end{tabular} & 21.201 & 14.946 & 54.297 \\  \midrule
\textbf{McCAD} & \textbf{19.901} & \textbf{14.847} & \textbf{42.364} \\ \bottomrule
\end{tabular}%
}
\caption{FID scores for ablations of adaptive FM and FA loss.}
\label{tab:perceptual_quality_results}
\end{table}

\textbf{Effect of feature components:}  The ablation of FM and FA loss components led to lower PSNR and SSIM scores in the synthetic contrasts, with only a marginal increase in PSNR observed in T1w synthesis as shown in Table \ref{tab:ablation_results}. Moreover, we have demonstrated the impact of these components on enhancing the perceptual quality of the synthetic outcomes. The results in  Table \ref{tab:perceptual_quality_results} show that adding the adaptive FM reduced the FID score of the synthesized results in many cases. The impact of the FA loss component was also notably significant in improving the perceptual quality.

\textbf{Effect of multi-contrast conditioning}: We also demonstrated the enhancement in performance achieved through multi-contrast imaging compared to single image-to-image contrast synthesis. The ablation results are presented in Table \ref{tab:multi-contrast-abl}, where multi-contrast synthesis offered higher accuracy in synthesis metrics.

\begin{table}[]
\centering
\resizebox{\columnwidth}{!}{%
\begin{tabular}{@{}ccccc@{}}
\toprule
 & \textbf{T1w  → FLAIR} & \textbf{T1w, T2w  → FLAIR} & \textbf{T1w → T2w} & \textbf{T1w, FLAIR → T2w} \\ \midrule
PSNR (dB) & 22.43±2.42 & \textbf{26.11±2.66} & 25.00±1.67 & \textbf{27.03±1.82} \\ \midrule
SSIM & 0.87±0.03 & \textbf{0.92±0.03} & 0.88±0.03 & \textbf{0.92±0.02} \\ \bottomrule
\end{tabular}%
}
\caption{Ablation study for multi-contrast MRI synthesis.}
\label{tab:multi-contrast-abl}
\end{table}

\begin{table}[]
\centering
\resizebox{0.7\columnwidth}{!}{%
\begin{tabular}{@{}ccc@{}}
\toprule
 & \textbf{McCaD - w/o adv. loss} & \textbf{McCaD} \\ \midrule

FLAIR, T2w → T1w & 38.5411 & \textbf{19.9018} \\ \midrule
T1w, T2w → T2w & 19.0530 & \textbf{14.8470} \\ \bottomrule
\end{tabular}%
}
\caption{FID scores for ablation of adversarial loss component}
\label{tab:adversarial-abl}
\end{table}

\textbf{Effect of adversarial loss component}: In contrast to conventional diffusion-based models lacking adversarial training, McCaD aims to improve the perceptual quality of the synthesized results. The results in Table \ref{tab:adversarial-abl} reveal that adding an adversarial loss component improved the perceptual quality of the results, as indicated by a lower FID score. In addition,   McCaD outperforms DDPM in computational efficiency by nearly 59 times speedup with the adaptation of adversarial-based training as shown in Table \ref{tab:inf-time-abl}.

\begin{table}[]
\centering
\resizebox{0.6\columnwidth}{!}{%
\begin{tabular}{@{}lll@{}}
\toprule
 & \textbf{DDPM} & \textbf{McCaD} \\ \midrule
Average inference time & 11690.93 & \textbf{199.3} \\ \bottomrule
\end{tabular}%
}
\caption{Time in milliseconds}
\label{tab:inf-time-abl}
\end{table}

% In addition,   McCaD outperforms conventional DDPM in computational efficiency, reducing inference time from 11,690 to 199 milliseconds—nearly 59 times speedup.

\section{ Conclusion}
% In this work, we have introduced McCaD, a multi-contrast MRI synthesis framework using an attention-guided, feature-adaptive adversarial diffusion model. McCaD employs multi-scale feature guidance from conditional contrasts during the denoising process, integrating precise adaptive feature maximization in the latent space. The proposed model achieves more accurate image synthesis compared with state-of-the-art methods for both healthy and tumor brain MRIs. We believe McCaD is a promising approach for synthesizing missing contrast in MRI.. We believe McCaD is a promising approach for synthesizing missing contrast in MRI.. We believe McCaD is a promising approach for synthesizing missing contrast in MRI.. We believe McCaD is a promising approach for synthesizing missing contrast in MRI.

In this work, we have introduced McCaD, a multi-contrast MRI synthesis framework using an attention-guided, feature-adaptive adversarial diffusion model. McCaD employs multi-scale feature guidance from the conditional contrasts in the denoising process with self-attention guidance by incorporating precise adaptive feature maximization in the latent space.  This ensures a more semantically guided denoising process by directing the model to focus on more attentive regions to improve the perceptual quality of results. The proposed model achieves more accurate image synthesis than SOTA methods for healthy and tumor brain MRIs. Additionally, we have evaluated our synthesis results on downstream task to further validate its clinical relevance. Therefore, we believe McCaD is a promising approach for multi-contrast MRI synthesis to enhance the efficiency and safety of the image acquisition process. 

While our proposed method has shown promising results, there is potential for further exploration in various tasks, including cross-modality translation (e.g., MRI to CT). Additionally, this study only considered three synthesis tasks and evaluating more critical synthesis tasks, such as contrast-enhanced MRI synthesis, will be important in clinical settings. Furthermore, McCaD is primarily designed for supervised training with co-registered datasets. Future studies could adapt it for unsupervised training by transforming our GAN-based method into a cycle-consistency architecture. Exploring different generative architectural frameworks could also be beneficial for improving the synthesized results.

%%%%%%%%% REFERENCES
% {\small
% \bibliographystyle{ieee_fullname}
% \bibliography{egbib}
% }

\end{document}